\documentclass[conference]{IEEEtran}
\IEEEoverridecommandlockouts
\usepackage{cite}
\usepackage{amsmath,amssymb,amsfonts}
\usepackage{algorithm}
\usepackage[noend]{algpseudocode}
\usepackage{nicefrac}
\usepackage{graphicx}
\usepackage{textcomp}
\usepackage{enumitem}

\usepackage[dvipsnames]{xcolor}
\def\BibTeX{{\rm B\kern-.05em{\sc i\kern-.025em b}\kern-.08em
    T\kern-.1667em\lower.7ex\hbox{E}\kern-.125emX}}
\usepackage{amsthm}
\usepackage{verbatim}
\usepackage{hyperref}

\newcommand{\unique}{\textsc{Unique}}

\newcommand{\segsum}{\textsc{Seg-Sum}}
\newcommand{\allgather}{\textsc{Ring-AllGather}}
\newcommand{\allreduce}{\textsc{Ring-AllReduce}}
\newcommand{\fillval}{\textsc{Fill-Values}}

\newcommand{\redHspace}{\hspace{-4mm}}
\newcommand{\redVspace}{\vspace{-4mm}}

\newcommand{\milind}[1] {{\color{blue}{Milind}}: {\color{red}{#1}}}

\newcommand{\ar}{\textsc{AllReduce}}
\newcommand{\ag}{\textsc{AllGather}}

\begin{document}

\title{
Language Modeling at Scale
}

\author{\IEEEauthorblockN{
Mostofa Patwary, Milind Chabbi, Heewoo Jun, \\ 
Jiaji Huang, Gregory Diamos, and Kenneth Church
}
\IEEEauthorblockA{
\textit{Silicon Valley AI Lab, Baidu Research} \\
Sunnyvale, California, USA \\
Corresponding Author: \texttt{patwarymostofa@baidu.com}
}
}

\maketitle

\begin{abstract}

We show how Zipf's Law can be used to scale up language modeling (LM) to take advantage of more training data and more GPUs.
LM plays a key role
in many important natural language applications such as speech recognition and machine translation.
Scaling up LM is important since it is widely accepted by the community that there is no data like more data.
Eventually, we would like to train on terabytes (TBs) of text (trillions of words).
Modern training methods are far from this goal, because of various bottlenecks, 
especially memory (within GPUs) and communication (across GPUs).
This paper shows how 
Zipf's Law can address these bottlenecks by grouping parameters for common words and character sequences, because $U \ll N$, where $U$ is the number of unique words (types) and $N$ is the size of the training set (tokens).
For a local batch size $K$ with $G$ GPUs and a $D$-dimension embedding matrix, we reduce the original per-GPU memory and communication asymptotic complexity from $\Theta(GKD)$ to $\Theta(GK + UD)$.
Empirically, we find $U \propto (GK)^{0.64}$ on four publicly available large datasets.
When we scale up the number of GPUs to 64, a factor of 8, training time speeds up by factors up to 6.7$\times$ (for character LMs) and 6.3$\times$ (for word LMs) with negligible loss of accuracy.
Our weak scaling  on 192 GPUs on the Tieba dataset 
shows a 35\% improvement in LM prediction accuracy by training on 93 GB of data (2.5$\times$ larger than publicly available SOTA dataset), but taking only 1.25$\times$ increase in training time,
compared to 3 GB of the same dataset running on 6 GPUs.

\end{abstract}

\section{Introduction} 
\label{sec:intro}

\begin{figure}[t!]
\begin{center}
  \includegraphics[width=\linewidth]{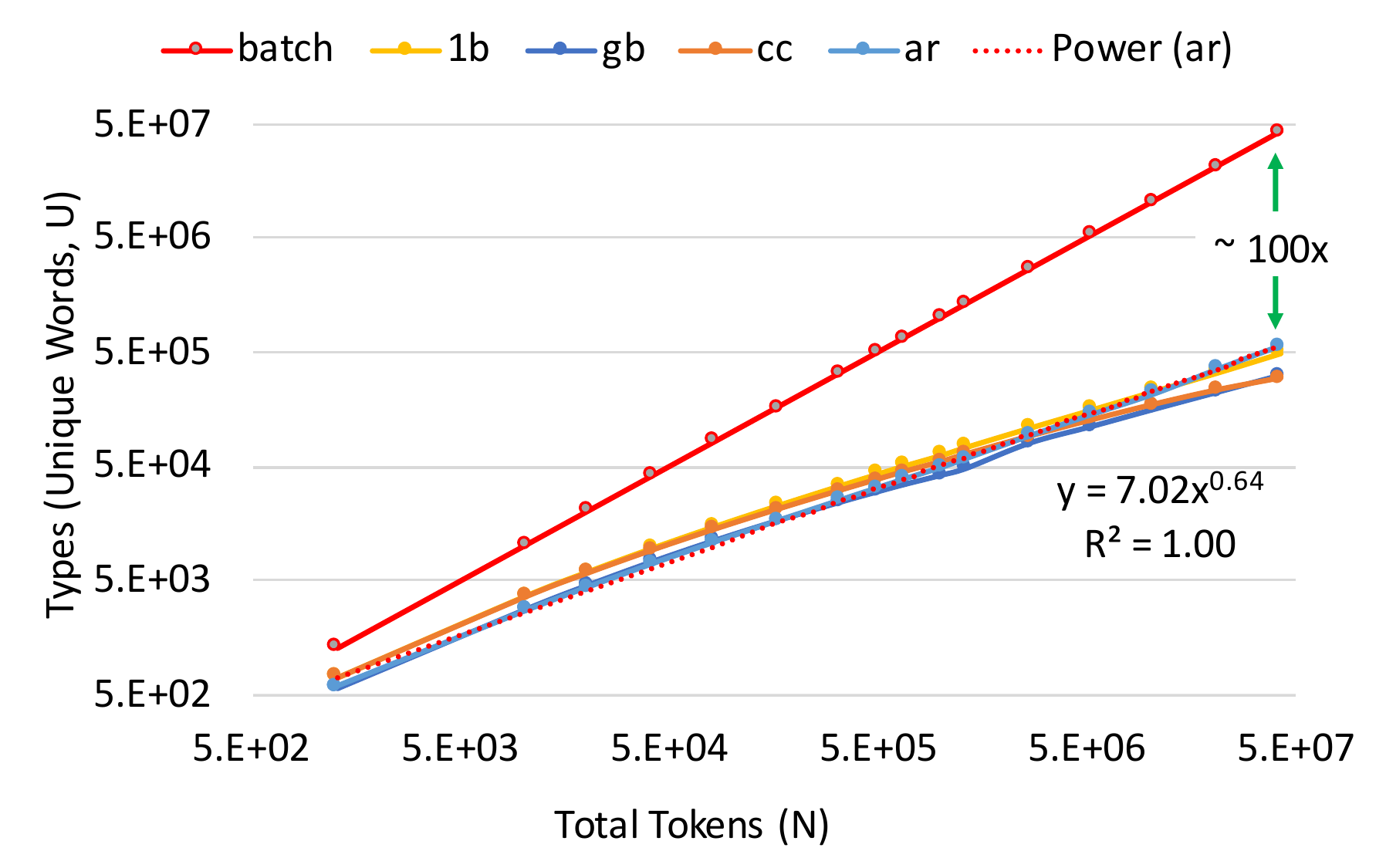}
\redVspace
  \caption{There are many more words (tokens) than unique words (types).  The red line is a baseline for $x=y$.  The gap between the data and the red line is large (even on $log$-scales), and increases as we scale up batch sizes and GPUs.
  The 4 lines
  correspond to 4 datasets: one Billion word ($1b$), Gutenberg ($gb$), Common Crawl ($cc$), and Amazon Review ($ar$).
  }
  \label{fig:statistics}
\end{center}  
\redVspace
\end{figure}

This paper will show how Zipf's Law~\cite{DBLP:books/aw/Zipf49, zipf:online, georgezipf:journal} can be used to help scale up language modeling (LM) to take advantage of more training data and more GPUs. Zipf's law is known to hold across 
many languages and wide variety of data sets~\cite{yu2018zipf, Isabel:2016}.
Zipf's law makes it clear that there are many more tokens than types, as illustrated in Figure~\ref{fig:statistics}.
It is common in language modeling to distinguish types (unique words) from tokens (non-unique words).
For example, the phrase, ``to be or not to be,'' consists of four types and six tokens.

In general, the number of \emph{unique} words in a training step is significantly smaller than the total number of tokens (the per-GPU batch size times the number of GPUs) and grows as a power law. 
Figure~\ref{fig:statistics} shows the number of types (unique words, $U$) on the y-axis as a function of tokens (non-unique words, $N$) along the x-axis.  The figure shows four datasets: 1-Billion word~\cite{chelba2013one} (1b), Gutenberg~\cite{ProjectGutenberg2018} (gb), Common crawl~\cite{Buck-commoncrawl} (cc), and Amazon review~\cite{McAuley:2015:IRS:2766462.2767755} (ar).  All four lines
fall well below the red line ($x=y$), labeled \emph{batch}.  This gap indicates an important opportunity for improvement.  The data fit a power law: $U \propto N ^{0.64}$.
When $N$ is 40-million total tokens in a training step, the number of unique words, $U$, is $\sim\!\!\!100\times$ smaller; and the gap continues to grow with $N$.

\begin{table}[t!]
\scriptsize
\caption{Datasets}
\begin{center}
\label{tab:datasets_stats}
\begin{tabular}{c|c|c|c|c} \hline
Datasets & \# Characters& \# Words & Bytes & Language \\  \hline
1-Billion Word\cite{chelba2013one} (1b)&	4.19B	&	0.78B	&	3.94GB	& English \\ 
Gutenberg\cite{ProjectGutenberg2018} (gb)	&	8.90B	&	1.81B	&	8.29GB	& English \\ 
Amazon Review\cite{McAuley:2015:IRS:2766462.2767755} (ar)	&38.76B	&7.01B&	37.04GB & English \\
Tieba\cite{baidutieba2018}	&	34.36B	&	NA	&	93.12GB  & Chinese \\ \hline
\end{tabular}
\end{center}
\redVspace
\end{table}

Language modeling is a fundamental task in natural language processing (NLP) and language understanding. 
It  predicts the next token (e.g. words, sub-words, or
characters) given the context (a sequence of surrounding tokens). 
Language modeling plays an important role in so-called noisy channel applications such as speech
recognition, OCR and spelling correction \cite{Church:1993}.  The noisy channel was introduced by
Shannon \cite{Shannon:1948,Shannon:1951}, and continues to be used in a number of more recent applications
such as:
natural language generation \cite{DBLP:journals/corr/abs-1708-01009}, 
machine translation \cite{DBLP:journals/corr/LuongPM15},
speech recognition \cite{DBLP:journals/corr/AmodeiABCCCCCCD15}, 
and text summarization \cite{DBLP:journals/corr/RushCW15}, to name a few. 
In the rest of this paper we use the abbreviation LM to mean Language Modeling or Language Model, which will be obvious from the context.

There is no data like more data. More data (and larger models) produce better estimates of sentence probabilities. 
Recent techniques leverage such large corpora by pre-training a neural language model and using the learned hidden representations to fine-tune on various NLP tasks. This simple but highly effective approach has achieved state-of-the-art results across many natural language understanding tasks
that have benefited from domain expertise and specialized architectures \cite{peters:2018:elmo,radford:2018:lm-pre-training}.


Unfortunately, more data and larger models also increase the training time~\cite{banko2001scaling,2018arXiv180801371P}. 
It is therefore of significant interest to accelerate the training
time of language modeling, specially by scaling the models to take advantage of the compute capability of high performance computing (HPC) resources such as GPUs. 
Although there have been several recent efforts to 
scale deep learning models in computer vision applications
\cite{DBLP:journals/corr/abs-1711-04325, DBLP:journals/corr/abs-1709-05011, DBLP:journals/corr/GoyalDGNWKTJH17},
less has been written on 
scaling language models and natural language
processing applications.
There are a couple of recent papers that scale LM implementations to a small number of GPUs~\cite{DBLP:journals/corr/abs-1804-03235, DBLP:journals/corr/abs-1803-08240}.  If we are going to scale up to terabytes, we will need to find a way to scale up to take advantage of many more GPUs.  This
work presents an important step in that direction.  


Scaling is challenging because the vocabulary (number of types) is large, and the training corpus (number of tokens) is even larger.  Modern neural network
based methods make use of word or character embeddings that tend to be large enough to run into memory and communication bottlenecks.
Unlike vision-related application, which employ an \ar{} over the gradients on all GPUs to update the local parameters on each GPU, LM-based applications cannot employ \ar{} due to the \emph{word/char embeddings}.
Instead, NLP applications use \ag{} operation over the embedding gradients, which results in memory demands and communication volume to grow proportional to the \emph{product of the number of GPUs and the batch size per GPU}. We elaborate more on the challenges in Section~\ref{sec:chall-techn}.

Prior work on scaling LMs tend to simplify the problem by limiting the size of the vocabulary, or
limited the number of GPUs.  For example, \cite{DBLP:journals/corr/abs-1804-03235} limited
the vocabulary to just $\sim\!\!\!24K$ words, a small fraction of the words in the corpus,
a large common crawl dataset~\cite{Buck-commoncrawl}.
Another example, \cite{DBLP:journals/corr/abs-1803-08240}, uses a large vocabulary, $\sim\!\!\!260K$,
but only four GPUs. The most recent study on
large scale language modeling  \cite{2018arXiv180801371P} demonstrates a scaling of up to 128 GPUs
but considers only character language models, where the vocabulary is tiny ($\sim\!\!\!100$).

This work will introduce three optimizers for scaling up:

\begin{enumerate}
    \item \emph{Uniqueness}: There are many fewer types than tokens ($U \ll N$) because of Zipf's law.  This observation allows us to turn a large, expensive \ag{} operation, employed in the \emph{input} word embedding layer, into a small \ag{} followed by an \ar{} operation, which changes the asymptotic complexity of memory and communication needed for updating gradients. 
    \item \emph{Seeding}: The so-called \emph{sampled} softmax\cite{DBLP:journals/corr/abs-1708-02182} employed in LMs to reduce the computational demands renders the {\it uniqueness} technique useless in LM's \emph{output} word embedding layer,
    because each GPU chooses a random subset of words, disobeying the word-frequency distribution.
We enforce a controlled randomization that obeys the power-law of word frequency distribution, which allows us to reap the benefits of {\it uniqueness} in the output word embedding layer.
    \item \emph{Compression}: Finally, we employ half-precision floating-point (FP16) numbers for data used in communication to further reduce bandwidth demands. FP16 reduces the communication volume by 50\%. 
    We recover the accuracy loss due to the lower precision via {\it compression-scaling}.
\end{enumerate}

\emph{Uniqueness} and \emph{seeding} reduce the asymptotic bounds of both communication volume and GPU memory size. \emph{Compression} reduces the communication volume by a constant factor.
We evaluate our optimizations on four large datasets (three publicly available and one internal).
Experimental evaluation demonstrates significant reduction in memory (within a GPU) and communication (across GPUs).
Our technique shows 8.6$\times$ memory reduction, which leads to 6.3X speedup for word LMs.
We demonstrate 6.7$\times$ (character LM) and 6.3$\times$ (word LM) speedup by scaling to 64 GPUs (8$\times$ more) with negligible loss of accuracy. 
Finally, we demonstrate weak scaling on Baidu {\it Tieba}\footnote{\url{https://tieba.baidu.com}} Chinese corpus (internal).
This paper will use a relatively small sample of what's available.
But even so, the sample of 93 GB we use is large enough to raise interesting scaling challenges:
2.5$\times$ larger than the publicly available state-of-the-art dataset.
Compared to a 3GB of the same dataset using 6 GPUs, when we scale to 32$\times$ more 
GPUs and data (192 GPUs and 93 GB, respectively), the running time increases by only 1.25$\times$, 
but provides an accuracy improvement of 35\%.

The paper is organized as follows. Section \ref{sec:chall-techn} describes the LM scaling challenges; Section \ref{sec:methodology} describes our techniques for scaling LM. Section \ref{sec:experiments} and 
\ref{sec:results_analysis} provide experimental setting and empirical results, respectively.
Section \ref{sec:related_works} discusses the related works, and Section \ref{sec:conclusions} concludes the paper.

\section{Background: Challenges in Scaling LM} 
\label{sec:chall-techn}


In this section we overview the state-of-the-art workflow for RNN-based language modeling.
Figure~\ref{fig:schematic} represents an anecdotal RNN-based language model akin to Bengio et al.~\cite{bengio2003neural}.
It consists of an input embedding, several feed-forward or recurrent (i.e. RNN) layers, an output embedding, followed by a softmax classifier layer.

\subsection{Language Model Basics}
LMs employ dictionary of commonly used terms.
For example, all letters (alphabets, numbers, punctuation) in a language forms the vocabulary for a \emph{character} LM, whereas all words in the dictionary form the vocabulary for a \emph{word} LM. 
A ``word'' is a unique entry in the vocabulary and a ``token'' is an instantiation of a word in a training set.

Assume a vocabulary $V$ of $|V|$ words.
Given a sequence of $K$ training tokens $w_1,w_2,w_3, \cdots, w_K$, where each $w_i \in V$, one can naively produce a $K \times |V|$ activation matrix $A$ as an input to RNN layers.
In this matrix, if $i^{th}$ input token is the $j^{th}$ word in $V$, $A[i][j]$ will be set to 1.
Such matrix will be extremely large for a large vocabulary, filled largely with zeros, and computationally very expensive for the subsequent layers of the neural network.

\begin{figure}[t!]
\centering
\includegraphics[width=\linewidth]{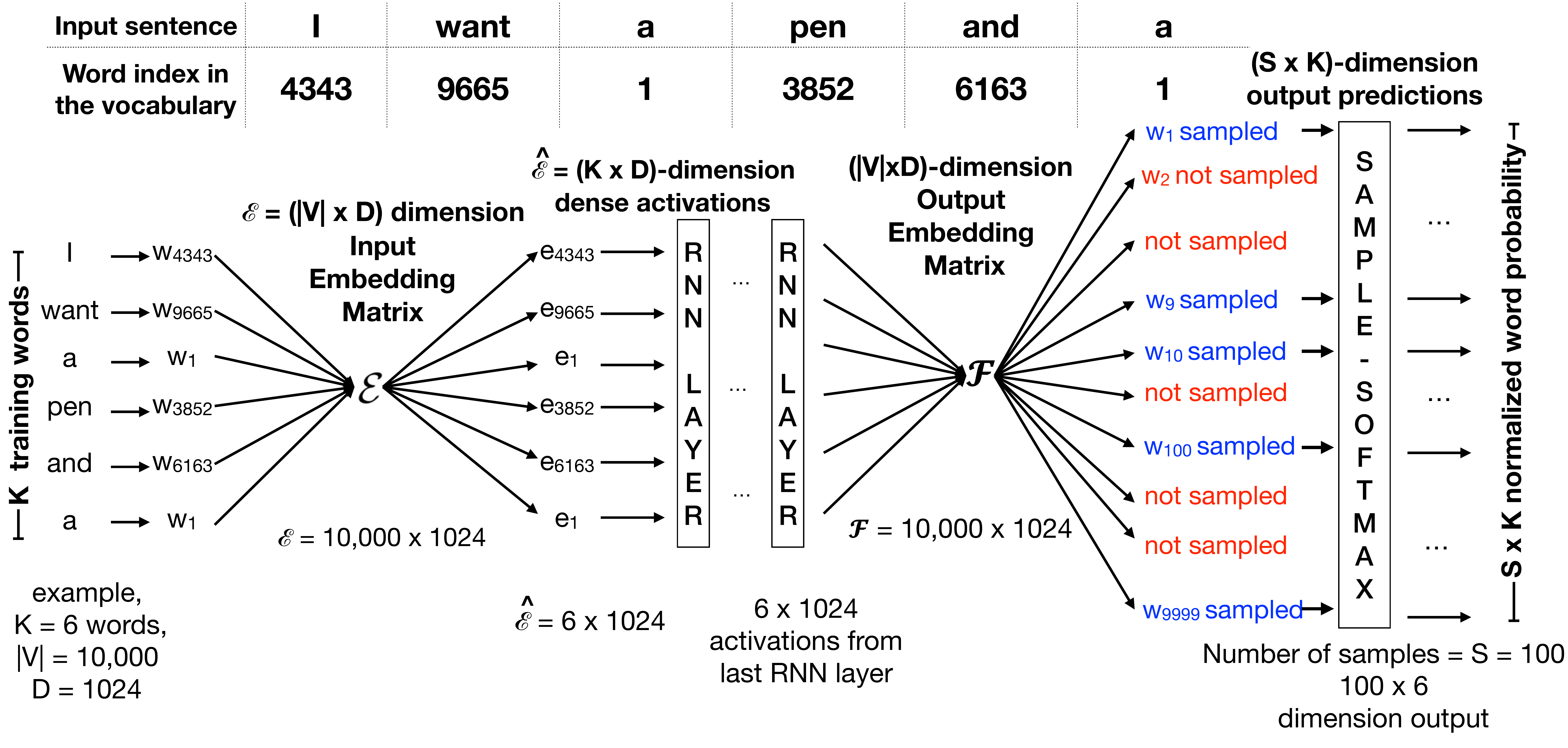}
\caption{{A schematic diagram of a typical RNN-based LM for a vocabulary $V$ and embedding dimension $D$. A $|V| \times D$ embedding matrix projects the input $K$ token sequence into a dense $K \times D$ matrix as the activations for the first RNN layer. After passing through the various number of RNN layers, a $|V| \times D$ output embedding matrix projects the prediction of each word in the vocabulary. For reducing the computational complexity, only a small number $S \ll |V|$ of words of the vocabulary are sampled and their normalized probabilities are computed by the sampled softmax layer.}}
\label{fig:schematic}
\end{figure}

LMs employ an ``embedding layer'' to reduce this size of activation input to the neural network.
Different words with related sentiments produce similar \emph{embedding vectors}, which are indistinguishable by the RNN layers.
The \emph{input} embedding layer projects the large, sparse input sequence of tokens into a small, dense  matrix $\widehat{\cal E}$.
To obtain $\widehat{\cal E}$, the model simply hash maps every input token $w_i$ to a D-dimensional vector $e_{w_i}$ of real numbers, where $D \ll |V|$, and produces a $K \times D$ dimension matrix.
The mapping uses a $|V|\times D$ embedding matrix $\cal{E}$.
The real numbers forming the embedding matrix $\cal{E}$ are learned during the training process.

Figure~\ref{fig:schematic} exemplifies this process. 
The input is a six-token sequence ``I want a pen and a'', where the word-index of each token is shown at the top of the figure. The first token ``I'' is $4343^{th}$ word in the vocabulary.
The token ``a'' appears twice, once at position three and again at position six, which becomes important during the back-propagation.
The first row of the dense, activation matrix $\widehat{\cal{E}}$ will be the $4343^{th}$ row from the embedding matrix  ${\cal{E}}$ corresponding to the token ``I'' at word-index $4343$.
The third and sixth rows of the activation matrix will be the $1^{st}$ row from the embedding matrix corresponding to the token ``a'' at word-index $1$, and so on.

During the back propagation of training, a gradient matrix $\Delta$ of dimension $K \times D$ is generated for the embedding layer.
Since, the embedding matrix $\cal{E}$ is $|V| \times D$ in dimension, a reverse mapping is performed from the $i^{th}$ row of the gradient matrix to the $j^{th}$ row of the embedding matrix. 
\emph{Since multiple rows of $\Delta$ may map on to the same row of $\cal E$, an updates to $\cal E$ is an accumulation operation.}

The RNN neural network consumes $\widehat{\cal{E}}$ and produces an intermediate representation of the input. 
The output of the last RNN layer is fed to an ``output embedding'' layer, 
which maps hidden states back to words, using inverse role of the input embedding layer.
The output embedding is a fully-connected layer that projects a lower dimension data back to the number of words in the vocabulary, so that the probability of every word can be predicted.
The softmax layer following the output embedding layer, produces a normalized probability distribution over \emph{all} words in the vocabulary. 
The probability of a word $w$ at a time step $t$ is calculated as $p(w^t | w^{<t})= \nicefrac{\exp(o_{w}^{t})}{\sum_{v\in V} \exp(o_{v}^t)}$, where $o_w^t$ is the output score from the last layer for the word $w$ at $t$. The softmax normalizes the output scores into a probability distribution.

The softmax calculation is computationally most expensive because the denominator is computed over all words in the vocabulary. 
Typical implementations reduce the computational complexity with various techniques, the simplest (and yet effective) is sampled softmax~\cite{chen2015strategies}, which computes the probability over a smaller, random subset over $V$. 
The sampled softmax is facilitated by making the output embedding choose a subset, e.g., 1\% of the words, in the entire vocabulary; typically, the words in the input are additionally included.

Because of the sampling, during back propagation, the gradients coming from the softmax later do not match the dimensionality of the output embedding layer.
Hence, the gradients are mapped back to the set of randomly chosen words during the forward pass, which is functionally similar to the back-propagation performed in the input embedding.
The uniform randomness does not ensure uniqueness of the chosen set of words in the output embedding.

\subsection{Parallelism in Language Models}
We now divert our attention to parallelizing the training process.
Data parallelism is the most common form of parallelism in neural networks; each processing entity (GPU in our case) holds the model but works on different $K$ input token sequence, drawn randomly from the entire training corpus. 
In fact, each GPU also consumes $K/c$ number of input sequences, where each sequence is of length $c$, and processes them in parallel; for brevity we refer to the entire data fed to a GPU as the \emph{local batch size} and represent its size with the symbol $K$.
While the forward propagation through the model can proceed unsynchronized across all GPUs, the gradient updates in each layer following the backward propagation needs to synchronize with all GPUs. 
The synchronization ensures that the model parameters on all GPUs are the same during the next training step. The so-called asynchronous  gradient update is an active research area and out-side the scope of our work.

To update the RNN parameters, the models perform an \ar{}~\cite{all-reduce-2018} to accumulate the gradients from all GPUs.
The accumulated gradients are used in updating the local weights.
The communication is over large gradient matrices (e.g. LSTM layers) and hence bandwidth bound; efficient implementations use a ring all-reduce technique~\cite{baidu-ring-allreduce2017}.
The input and output embedding connections are special and pose additional challenges.

\begin{figure}[t!]
\centering
\includegraphics[width=.8\linewidth]{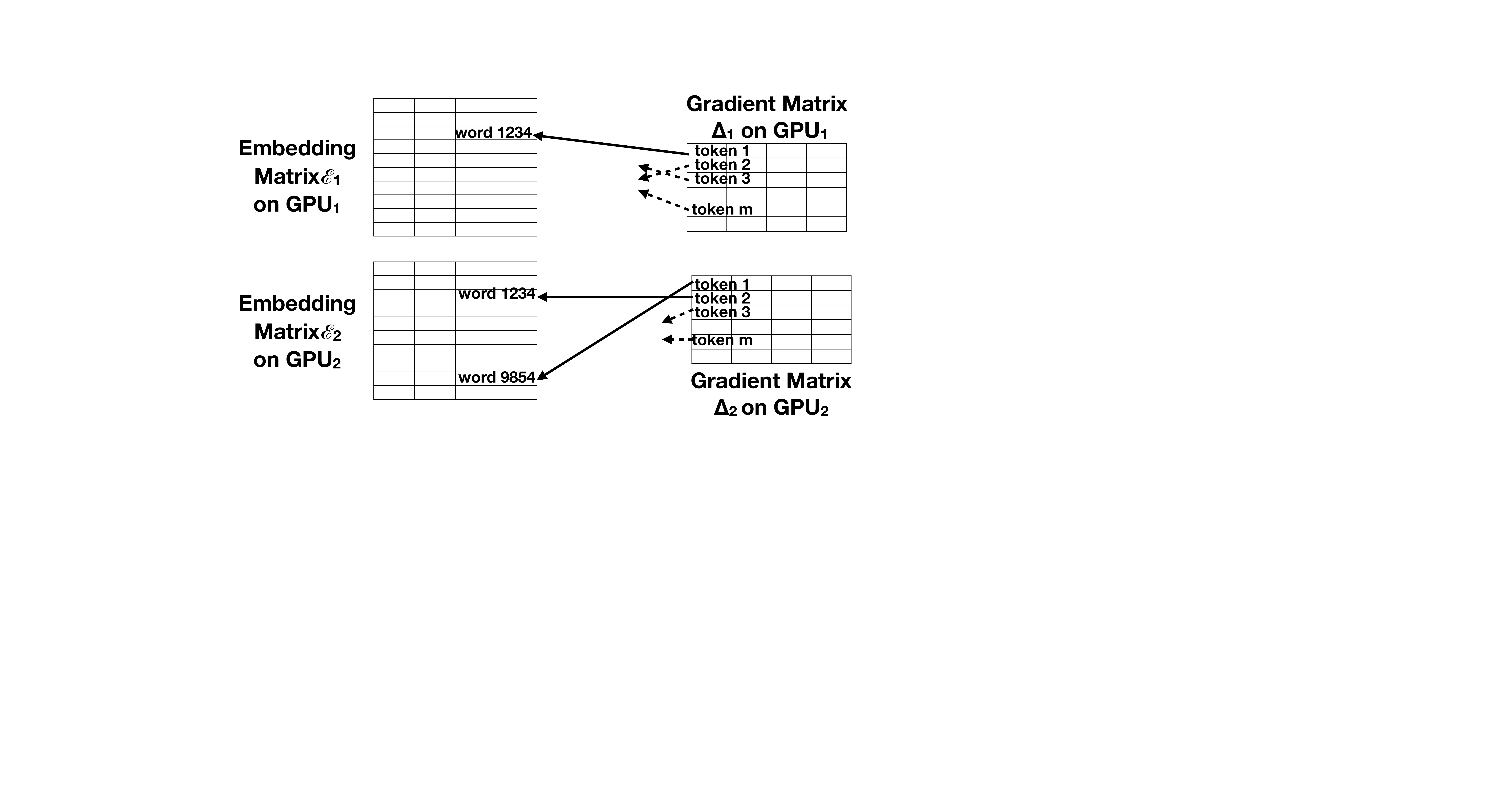}
\caption{GPU$_1$ and GPU$_2$ independently compute their local gradients $\Delta_1$ and $\Delta_2$, respectively. However, the updates to their embedding matrices ${\cal E}_1$ and ${\cal E}_2$ needs to be synchronized. The first token (row) of $\Delta_1$ maps to $1234^{th}$ word (row) in ${\cal E}_1$ and the first token in $\Delta_2$ maps to  $9854^{th}$ word in ${\cal E}_2$. As result, one cannot accumulate gradients with an \ar{} operation. LM implementations perform the space- and communication-expensive \ag{} to circumvent the problem.}
\label{fig:multigpuindex}
\redVspace
\end{figure}

During the same time step, each GPU $i$ can have its own $K$ training tokens: $w_{i1}, w_{i2}, \cdot, w_{iK}$, different from the $K$ training tokens on another GPU  $j$ represented by $w_{j1}, w_{j2}, \cdot, w_{jK}$, which is the reason for complication in the embedding layers.
If the $p^{th}$ tokens is not an instantiation of the same word on the two GPUs, (that is, $w_{ip} \ne w_{jp}$), which is often the case, then the gradients  computed for the $p^{th}$ tokens ($\Delta_{ip}$ and $\Delta_{jp}$) on two different GPUs do not map to the same row of the embedding matrix during the reverse mapping step.
This is depicted in Figure~\ref{fig:multigpuindex}, where the gradient for the first tokens on GPU$_1$ maps to the $1234^{th}$ row of the embedding matrix, whereas the gradient for the first tokens on GPU$_2$ maps to the $9845^{th}$ row of the embedding matrix.
Furthermore, since the words need not be unique across the GPUs, the gradient for the second tokens on GPU$_2$ maps to the $1234^{th}$ row of the embedding matrix.
We remind that the two embedding matrices ${\cal E}_1$ and ${\cal E}_2$ must remain the same across updates.

Since gradients at the same index on two different GPUs may map to two different rows of the embedding matrix, one cannot perform an \ar{} operation over all $\Delta_i = K \times D$-dimensional dense gradients.
State-of-the-art implementations, hence,  perform an \ag{}, which collects all $K \times D$ matrices from all $G$ GPUs ($G-1$ other GPUs to be precise) and then applies the gradients to the local embedding matrix.
The \ag{} operation requires $\Theta(G \times K \times D)$ local memory to hold $G$ number of $\Delta$ matrices, and the communication time is also bounded by $\Theta(G \times K  \times D)$.
Finally, the time to  update each ${\cal E}_i$ is also bounded by the $\Theta(G \times K  \times D)$. 
Not all $G \times K$ words are unique; words can repeat within a token sequence both on the same GPU and on different GPUs.
Hence, while concurrently updating different rows of $\cal{E}$ using the parallelism on GPUs, \emph{the rows under update are locked to prevent races.}
Such locking is necessary even in the single GPU case since the words can repeat within a sequence presented to the same GPU.

The updates to the output embedding is analogous to the input embedding in the presence of sampled softmax due to random, sparse word selection.
If each GPU computes the probability of $S$ randomly chosen output words, during the gradient update, it has to gather the updates from all other $G$ GPUs and then update the local output embedding matrix.
The number of samples is proportional to the local batch size, that is $ S \propto K$.
As before, implementations perform an \ag{} to accomplish this task.
If the output embedding is a vector of size $D$,  the \ag{} operation requires $\Theta(G \times K \times D)$ local memory to hold the entire update; the communication and local update time are bounded by $\Theta(G \times K \times D)$. Implementations may use different dimensions for input and output embeddings, but it is less common.

In summary, embedding layers are the performance limiters in LM implementations.
LMs' local memory footprint grows proportional to the product of local batch size and the number of GPUs ($\Theta(G \times K)$).
Since, GPUs have a limited memory ($\sim$16GB), one cannot scale LMs beyond a handful of GPUs.
LM's communication volume and GPU memory footprint grow proportional to the number of GPUs times the local batch size.
Thus, large-scale language modeling (whether using a large batch size or a large number of GPUs or both) becomes communication bound, runs out of memory, and consequently fails to scale beyond a few GPUs or suffers from poor parallel efficiency; Section~\ref{sec:results_analysis} provides empirical data in this regard.

\section{Methodology: Scalable Language Modeling} 
\label{sec:methodology}

We, now, describe how we overcome the fundamental limiting factors in scaling LMs.
Although, at the outset, the algorithmic complexities seem to limit scalability, studying the word distribution in a training corpus offers optimization insights.
Word distribution empirically follows the well-known Zipf's  power law \cite{DBLP:books/aw/Zipf49, zipf:online, georgezipf:journal}: ``given some corpus of natural language utterances, the frequency of any word is inversely proportional to its rank in the frequency table. Thus, the most frequent word will occur approximately twice as often as the second most frequent word, three times as often as the third most frequent word''.
We exploit this domain knowledge on the word distribution to reduce the previously shown complexity bounds on scalability.
The larger the batch size or more the number of GPUs, higher the opportunity to exploit the Zipf's law frequency distribution, discussed in ~\cite{yu2018zipf, Isabel:2016}.

The rest of this section describes our strategy exploiting this observation for achieving better scalability.
We first explain its application to the input embedding layer.
We then describe an additional optimization---controlled seeding---to make scheme applicable to the output embedding layer.
We end the section with an orthogonal optimization, half-precision communication, which provides
an additional improvement in scaling.

\begin{figure}[t!]
\centering
\includegraphics[width=\linewidth]{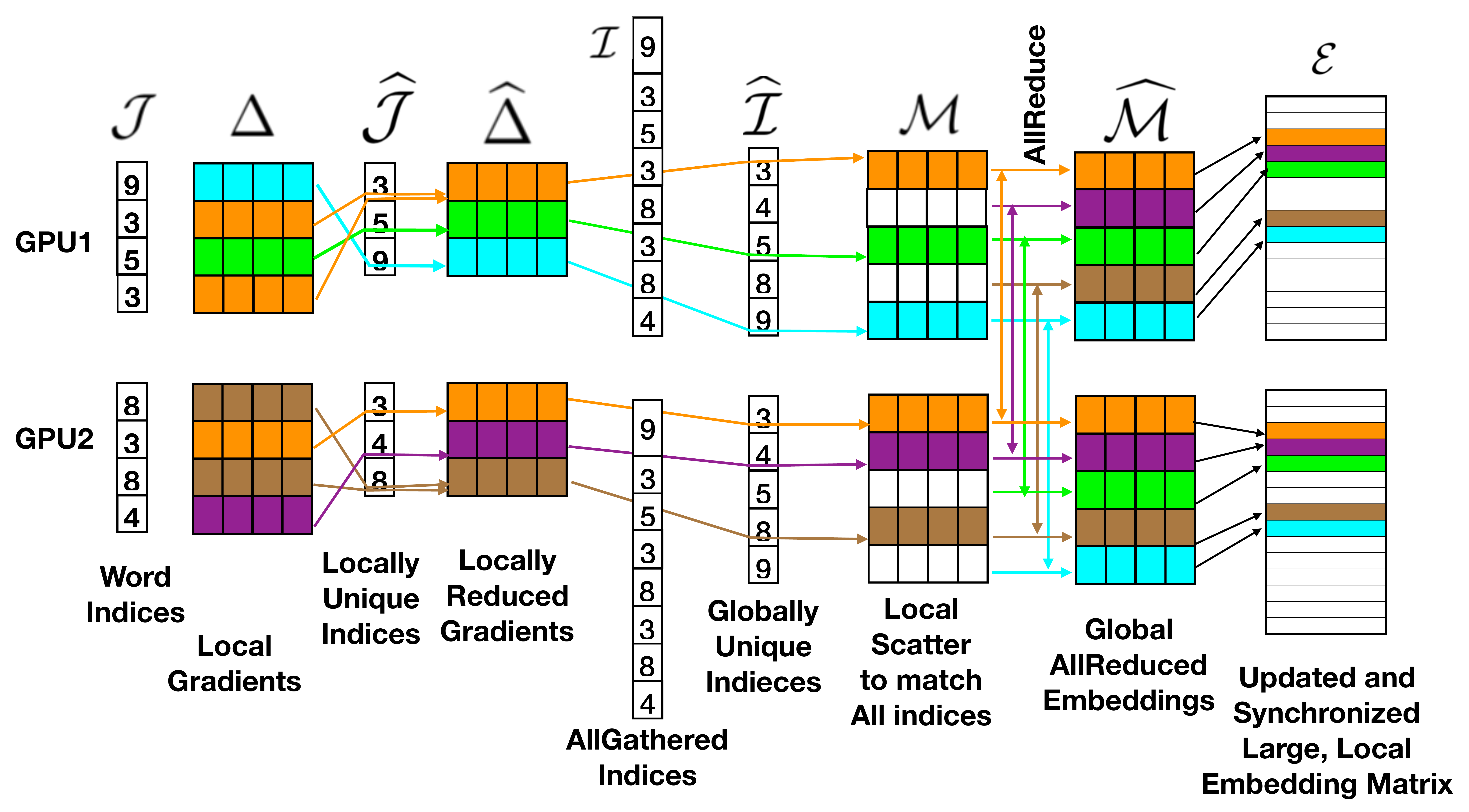}
\caption{{A demonstration the \unique\ approach in updating each GPU's embedding matrix $\cal{E}$ from its gradients $\Delta$. Expensive \ag{} over the entire gradients is converted into a sequence of local unique index computation ($\cal{J} \rightarrow {\cal{J}}$), all-gather over indices ($\widehat{\cal{J}} \rightarrow \cal {I} \rightarrow \widehat{\cal I}$), local scatter of gradients ($\Delta \rightarrow \cal{M}$)  and an \ar{} to produce $\widehat{\cal{M}}$, which is used to update $\cal E$.}}
\label{fig:methodology}
\redVspace
\end{figure}

\subsection{Exploiting word uniqueness to reduce communication and memory demands of embedding layer.} 
\label{subsec:exploit-uniquness}

Figure~\ref{fig:methodology} depicts our strategy. To give a high-level intuition, we perform an \ag{} over the \emph{word indices} to know all unique words presents in a training step. 
Then, each GPU re-arrange its local gradients into a matrix such that a gradient vector corresponding to a given word appears at the same position (row) across all GPUs. We then perform an \ar{} over the re-organized gradients.

Let the local batch of $K$ tokens on GPU $i$ contain $U_i \le K$ unique words.
Let the $K$-dimension vector $\cal{J}$ on each GPU hold the word index corresponding to each token in its input sequence.
Our strategy can be described in the following  sequential steps.

\begin{enumerate}[leftmargin=*]
\item On each GPU, compute the vector $\widehat{\cal{J}}$, which holds the word indices of only \emph{unique} words in its input sequence. 
In others words, $\widehat{\cal{J}}$ is a vector of ``types'' present on that GPU.
\item On each GPU  $i$, perform a \emph{local reduction} of the gradient vectors, so that the gradient vectors corresponding to the same words are accumulated into a single vector. 
Now, each GPU $i$ has a gradient matrix ${\widehat\Delta}_i$ of dimension $U_i \times D$.
\item Perform an \ag{} over ${\cal{J}}$ vectors from all GPUs. 
This \ag{} consumes only $\Theta(G \times K)$ memory as opposed to the traditional \ag{} that required $\Theta(G \times K \times D)$ memory.
Let the resulting vector be $\cal I$, which is same on all GPUs.
\item On each GPU, perform a \emph{local} filter operation over the $G \times K$ indices (vector $\cal I$) to extract all unique word indices to produce vector $\widehat{\cal I}$.
In other words, $\widehat{\cal I}$ holds all ``types'' in a training step.
Let the elements in the $\widehat{\cal I}$ be totally ordered and let us maintain a mapping from an entry in  ${\cal J}$ to the corresponding entry in $\widehat{\cal I}$ and transitively from $\widehat{\cal J}$ to $\widehat{\cal I}$, which are local operations.
Now, each GPU has a consistent view of all word indices present in this time step; if the $p^{th}$ entry of $\widehat{\cal I}$ on GPU $i$ points to $q^{th}$ row of its $\cal{E}$, so does the $p^{th}$ entry of $\widehat{\cal I}$ on another GPU $j$.
Let each GPU infer that in total there are $U_g$ unique words in this time step.
\item On each GPU $i$, expand the ${\widehat\Delta}_i$ matrix obtained in step 2 from a $U_i \times D$ matrix into a $U_g \times D$ matrix via a local scatter operation. 
The non existing entries are filled with zeros.
Let this expanded matrix be called ${\cal M}_i$.
Note that $U_i \le U_g \ll G \times K \ll  |V|$.
\item Perform an \ar{} over all ${\cal M}_i$, each of which is of the same $U_g \times D$ dimension.
This step has communication cost of $\Theta(U_g \times D)$. Let the resulting matrix be $\widehat{\cal M}$.
\item Update the local embedding matrix with the the values in $\widehat{\cal M}$ using $\widehat{\cal I}$ to map index in $\widehat{\cal M}$ with row in $\cal{E}$.
\end{enumerate}
The total space and communication complexities are:
$\Theta\big( (G \times K) + (U_g \times D) \big)$, which is a significant reduction from the original $\Theta( G \times K \times D )$.
Since $U_g \propto (G \times K)^{\alpha}$, we have reduced both time and memory complexity from  $\Theta( G \times K \times D )$ to  $\Theta\Big( \big(G \times K \big)^{\alpha} \big( (G \times K )^{1-\alpha} + D \big)\Big)$, where $\alpha$ is the exponent in Zipf's power-law in word frequency distribution.

Consider a real-word example, where the sequence length is $c=150$, the number of sequences per GPU is $128$, which makes a local batch size of $K=150*120=19,200$, and the embedding dimension is $1792$. 
In this setting, with 32-bit floating-point gradients, on 256 GPUs, the old scheme of \ag{} would require $35.2$ GB of memory per GPU.
However, with our {\it uniqueness} technique where the power-law exponent is $0.64$, we would require only $0.137$ GB
of memory per GPU---a $256\times$ memory saving. 

An additional benefit is that since all indices are unique when updating the local model in step 7, no two indices are simultaneously attempting to update the same embedding vector in $\cal{E}$ and hence no serialization bottleneck.
To better appreciate this fact, imagine that in a set of updates, if 50\% of the tokens are all the same highest-frequency word, the  updates to their corresponding embedding vector would be serialized wasting the available parallelism on a GPU.
This problem is eliminated in our update scheme that has no duplicate words.

\subsection{Controlled randomness to reduce communication and memory demands of softmax layer.}
\label{subsec:exploit-seeding}

The \emph{uniqueness} technique is not directly applicable when updating the \emph{output} embedding matrix in the presence of  sampled softmax because the sampling can choose different words on different GPUs.
For a large vocabulary, the probability of choosing the same word at the same index is minuscule; and the total words selected by all the GPUs grows proportional to the number of GPUs times the local batch size.
Thus, we lose the power-law distribution of words when updating the output embedding with the gradients.

An easy approach would be to force all GPUs to use the same random seed, so that they all choose the same set of random words in each time step.
Although, the same seed makes the updates to the output embedding amenable to the same optimization described in Section~\ref{subsec:exploit-uniquness}, the loss of randomness leads to loss of diversity, which results in poor learning and degrades accuracy.
Thus, there is a trade-off: each GPU with a different random seed has a good accuracy but poor scalability, whereas each GPU with the same random seed has a poor accuracy but good scalability.

Interestingly, the trade-off is not binary; there is a spectrum of choices to make.
Instead of all same seed or all different seeds, 
we make a subset of GPUs use the same seed.
We evaluated  the number of seeds equal to 
$\log_2$, $\log_e$, and $\log_{10}$ of the number of GPUs.
\emph{We empirically observed that the number of different seeds needed to produce accuracy matching all different seeds matches the power law.}
Meaning, with $G$ GPUs, we only need $G^{\alpha}$ unique random seeds to achieve a very good accuracy (empirically $\alpha=0.64$) while enjoying the benefits of few unique words and hence less communication and memory overhead. We present the details in Figure~\ref{fig:seeding-techniques} in Section~\ref{sec:results_analysis}.

Equipped with this technique, the rest of the procedure in updating the output embedding matrix is the same as that of the input embedding layer.
When $S$ is the number of sampled words per GPU, the total space and communication complexity of the updates performed in the output embedding layer are:
$\Theta\big( (G \times S) + (U_g \times D) \big)$, which is a significant reduction from the original $\Theta( G \times S \times D )$.
Since $U_g \propto (G \times S)^{0.64}$, in practice, we have reduced both time and memory complexity from  $\Theta( G \times S \times D )$ to  $\Theta\big((G \times S)^{0.64} \times D \big)$.


\subsection{Lower precision to reduce communication}
\label{subsec:compression}

Deep learning models are usually trained using 32-bit 
floating point (FP32) numbers. However, due to the increased gap between computation required vs. delivered \cite{compute-need-delivered-2017} 
for deep learning applications, 
reduced precision (e.g. 16-bit floating point numbers, FP16) is gaining popularity. 
Recently, \cite{micikevicius2017mixed, mixed-precision-2018} showcased
that FP16 based models can be trained with
negligible loss of accuracy. 
It uses a {\it loss-scaling} technique,
to minimize the number of gradient values 
becoming zeros, due to lower precision.
The idea is to multiply the training loss (e.g. cross-entropy) by a scaling factor, $F$ (e.g. 256, 512, and 1024) before computing gradients and then divide the gradients by $F$ before updating the weights. 
This method reduces the memory footprint by 50\% and works well on a wide range of applications including image recognition and machine translation \cite{micikevicius2017mixed}.

We use the same concept of lower precision to reduce communication among the GPUs. 
We  down-cast each FP32 tensor to FP16,
communicate, and 
up-cast the FP16 tensor to FP32
at the receiving end. This reduces the communication by 50\%. To minimize loss introduced by lower precision, we perform {\it compression-scaling}, that is, multiply the FP32 tensor by a scaling factor, $F$ before down-casting, and divide again by $F$ after up-casting. We call this method {\it compression}.

\section{Experimental Setup} 
\label{sec:experiments}


We performed all the experiments on a 50-node cluster.
The software and hardware configurations are tabulated below.

\begin{table}[h!]
\scriptsize
\begin{center}
\caption{System configuration.}
\redVspace
\label{tab:system}
\begin{tabular}{r|l} \hline
\# Nodes & 50 \\
Interconnect & Infiniband FDR @ 15GB/s bidirectional bandwidth\\
CPUs/node & 2 $\times$  20-core Intel Xeon E5-2660 v3 @2.6 GHz\\
Memory/node &  400GB DDR  \\
GPUs/node &  8 $\times$ GeForce GTX Titan X @ 32 GB/s PCIe  bidirectional b/w\\
GPU memory & 12GB HBM2 \\
peak FLOP/GPU & 6.1 TFLOP/s (32-bit floating point numbers) \\
Software & Tensorflow 1.4 \cite{tensorflow2015-whitepaper}, CUDA 8.0.61, CUDNN 6.0.20. \\
& cuda-aware OpenMPI 2.0.1 \\ \hline
\end{tabular}
\end{center}
\redVspace
\end{table}

We use one GPU per MPI process in all of our experiments.
Communication among the GPUs (both inter and intra-nodes) use cuda-aware MPI collectives incorporated 
in Tensorflow. 

\subsection{Datasets}

We used four datasets in our experiments, three in English and one in Chinese. 
One of them, the 1-Billion word \cite{chelba2013one}, is the commonly used one to perform 
language modeling experiments \cite{DBLP:journals/corr/JozefowiczVSSW16}. 
We used the Gutenberg \cite{ProjectGutenberg2018} dataset to better
understand that our techniques are dataset independent. We used the Amazon Reviews dataset \cite{McAuley:2015:IRS:2766462.2767755}, which was used in a recent scaling paper, \cite{2018arXiv180801371P}. We finally used a subset of an internal Chinese dataset curated from Baidu's internet forum called {\it Tieba} \cite{baidutieba2018} to perform a Hero scale run using
192 GPUs. 
To train the models and to test the accuracy, 
we split the the first two datasets into 99:1 ratio and the
last two into 1000:1 ratio (similar to \cite{2018arXiv180801371P}).
Each split is created by sampling without replacement and a fixed random seed.
The vocabulary for character language model includes all alphanumeric characters and common symbols (98 in
English and $\sim$15K in Chinese).
For word language models, we use the 100,000 most frequent words after lower-casing and tokenization\cite{bird2009nltk} as the vocabulary for each corpus. The number of unique words can range from 2M to 24M in the corpora we considered, but vocabularies created by this simple procedure account for 99\% of the text in each data set.
A summary of all the above datasets is presented in Table \ref{tab:datasets_stats}.

\subsection{Model Architectures}

To analyze scaling and accuracy, we take the character and word language models as test-cases 
for small and large vocabulary,
respectively. 
For word language model, we use the baseline LSTM based SOTA model from \cite{DBLP:journals/corr/JozefowiczVSSW16}. 
The model consists of one LSTM layer with 2048 cells. The projection dimension we used is 512. 
The batch size per GPU is 32 and sequence length is 20. This configuration with $\sim\!\!\!800K$ 
vocabulary (as used in \cite{DBLP:journals/corr/JozefowiczVSSW16}) requires more than 9.8 GB of memory 
for the model parameters and activations. We therefore used a reduced vocabulary size of $\sim100K$ so that 
required memory is much lower (1.3 GB) and also the CPU-GPU traffic reduces significantly. 
In the experiments, we used stochastic gradient descent (SGD) for optimizing per-sequence word cross-entropy loss using a sampled softmax layer, with 1024 random samples per GPU. 
The learning rate is $0.2 \times \log_e (|nodes|)$ with decay factor ranging from 0.85 to 0.95 in the experiments. 
In our experiments, each {\it node} consists of 8 GPUs.

For the character language model, we use the SOTA model similar to \cite{DBLP:journals/corr/abs-1712-00409}. The model
consists of a recurrent highway network (RHN) layer of depth 10, each with 1792 LSTM cells. The model consists of 213 million parameters.  
We use 128 batch size per GPU with sequence length of 150. 
We use {\it Adam} with weight decay and dropout for optimizing the character cross-entropy loss using a full softmax layer. We used a learning 
rate of $10^{-3} \times \log_e (|nodes|)$ with decay factor ranging from 0.85 to 0.95 in the experiments.

\section{Results and Analysis} 
\label{sec:results_analysis}

In this section, we present the experimental results obtained by our proposed methodology. We use 
word and character language model as test-cases for large and small vocabulary, respectively. 
We showcase accuracy and speedup comparison along with details analysis for 1-Billion and Gutenberg datasets using 16, 32 and 
64 GPUs.
We later present results
of a \emph{hero-scale} run on the Tieba dataset 
using 192 GPUs. Finally, we compare our results with existing works on the Amazon review dataset.

\begin{figure}[h!]
\begin{center}
  \includegraphics[width=0.45\textwidth]{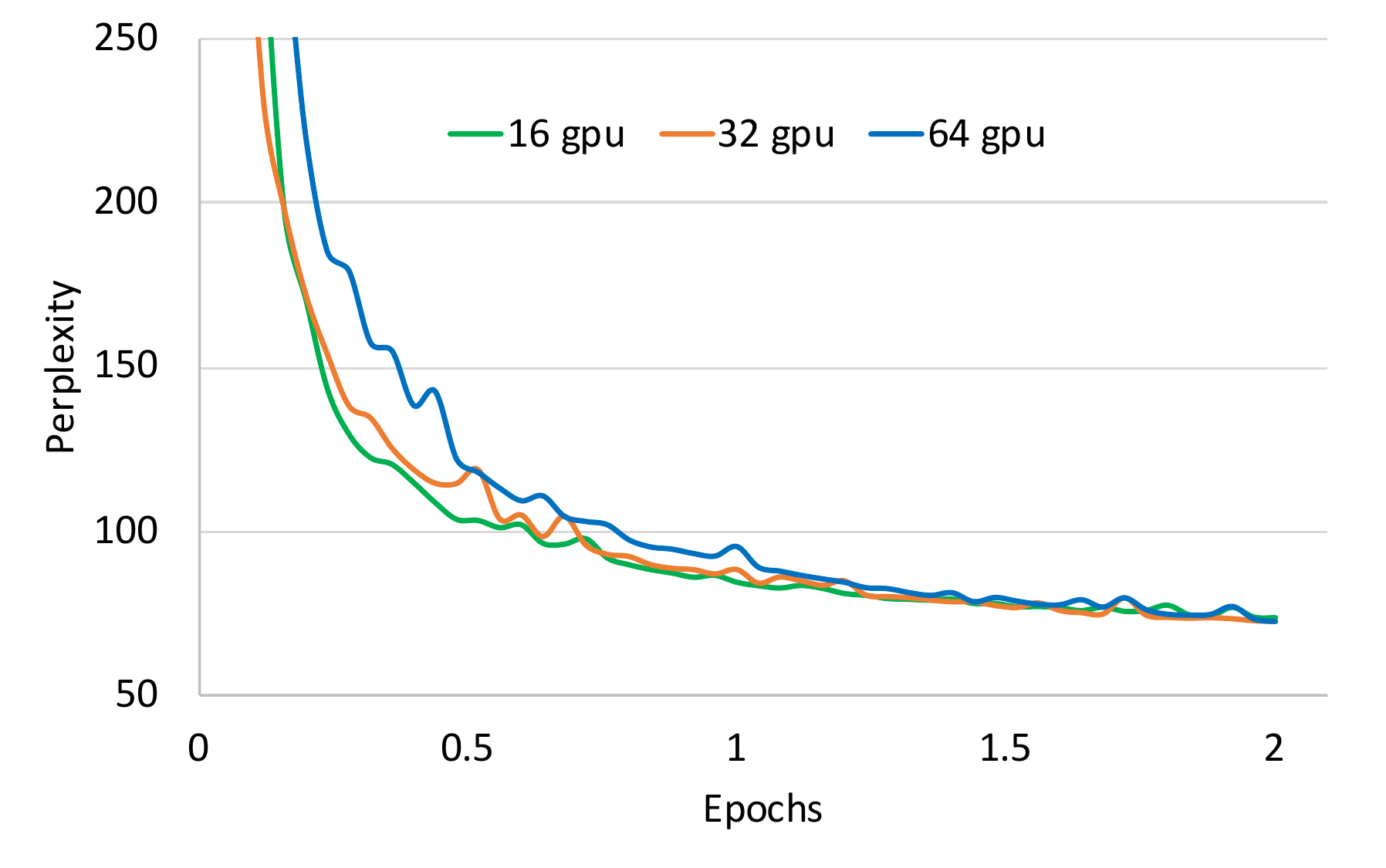}
  \redVspace
  \caption{Accuracy of word language model on the 1-Billion word dataset using 16, 32, and 64 GPUs.}
  \label{fig:wordLM-accuracy-gpus}
\end{center}  
\redVspace
\end{figure}

\subsection{Word Language Model}
\label{subsec:word-LM-results}

We first present the accuracy and speedup achieved by the
word language model with large vocabulary ($\sim100K$). 
We use three combinations of GPUs, 16, 32, and 64, to perform the scaling experiments
with batch size of 512, 1024, and 2048, respectively. The sequence length used was
20, therefore, per iteration the number of tokens (words) processed 
was 10240, 20480, and 40960, 
respectively for the three GPU combinations. 
We use perplexity (lower is better) to 
compare accuracy, which measures how well a model 
is capable to compute the probability distribution 
to predict words or characters. 

Figure \ref{fig:wordLM-accuracy-gpus} shows the accuracy validation perplexity up to 2 epochs 
for the 1-Billion dataset. 
The perplexity becomes indistinguishable with increasing epochs. 
For example, at Epoch 1, the perplexities are 84.3, 87.9, 	and 95.3 for 16, 32, and 64 GPUs. The values reduces to 73.5, 72.1, and	72.4, respectively at Epoch 2. We 
realized that 32 and 64 GPUs produce better perplexity 
compared to 16 GPUs run. The trend continues in the later epochs as well (e.g. 67.7, 63.7, and 63.6 at epoch 5).
We achieved similar trend with accuracy for the Gutenberg dataset. For example, we found perplexity of 76.7, 77.4 and 81.1 at epoch 1 whereas these values become 63.0, 63.6, and 67.1 at epoch 3 using 16, 32, and 64 GPUs respectively.
We use $0.2$ as the base learning rate 
(for 8 GPUs) and then used a multiplying factor of $\log_e |nodes|$ (e.g. $0.41$ for 64 GPUs) as we increase the number of GPUs.

\begin{table}[t!]
\scriptsize
\begin{center}
\caption{Per epoch time (hours) on Titan X GPUs for word LM using 1-Billion word dataset. 8-GPUs is the baseline for computing parallel efficiency.  $* =>$ out of GPU memory.}
\label{tab:epoch_time_word_gpus}
\begin{tabular}{c||c|c||c|c} \hline
\multicolumn{1}{c||}{} & \multicolumn{2}{c||}{Without Our Technique} & \multicolumn{2}{c}{With Our Technique} \\  \hline
GPUs & Time & Parallel & Time & Parallel \\ 
 & (hours) & Efficiency & (hours) & Efficiency \\ \hline 
8 &	35.1	& 100\% &	14.6 & 100\%	\\ 
16 &	41.1	& 43\% &	8.1 & 90\%	\\
24 &	40.4	& 29\% &	6.4 & 76\%	\\ 
32 	&	$*$	& - &	5.4 & 67\%	\\
64 	&	$*$	& - &	4.5 & 40\%	\\ \hline
\end{tabular}
\end{center}
\redVspace
\end{table}

Table \ref{tab:epoch_time_word_gpus} shows the time taken per epoch by the 
word language model for
1-Billion word dataset while varying the number of GPUs, keeping the 
local batch size fixed. Using our techniques, we found that per epoch time
using 8 GPUs is 14.6 hours. If we increase the number of GPUs by 8$\times$ (i.e. 64 GPUs), 
the training time reduces to 4.5 hours (3.2$\times$ speedup). Compared to the 8 GPUs run without our techniques, the
speedup becomes 7.7$\times$.
Without our techniques the code struggles to achieve parallel efficiency of 29\% using 24 GPUs and goes out of memory with more GPUs.
In contrast, our techniques deliver 76\% parallel efficiency using 24 GPUs. The value become 40\% when we use 64 GPUs 
using our approaches ($>$ 24 GPUs run without our techniques). We found similar results for the Gutenberg dataset 
($2.4\times$ speedup using 8$\times$ more GPUs and a parallel efficiency of 30\% on 64 GPUs). Compared to the 8 GPUs run without our techniques, the speedup becomes 6.3$\times$.
The lower speedup in word LM when compared to our own 8 GPUs run is due to the low computational intensity (136 GFLOP/iter) of word LMs; character LMs achieve higher speedup (2,721 GFLOP/iter) as shown in the next section.
We obtained 2.44 TFLOP/sec
(40\% of peak FLOPS) in the experiments.

\begin{figure}[t!]
\begin{center}
  \includegraphics[width=0.45\textwidth]{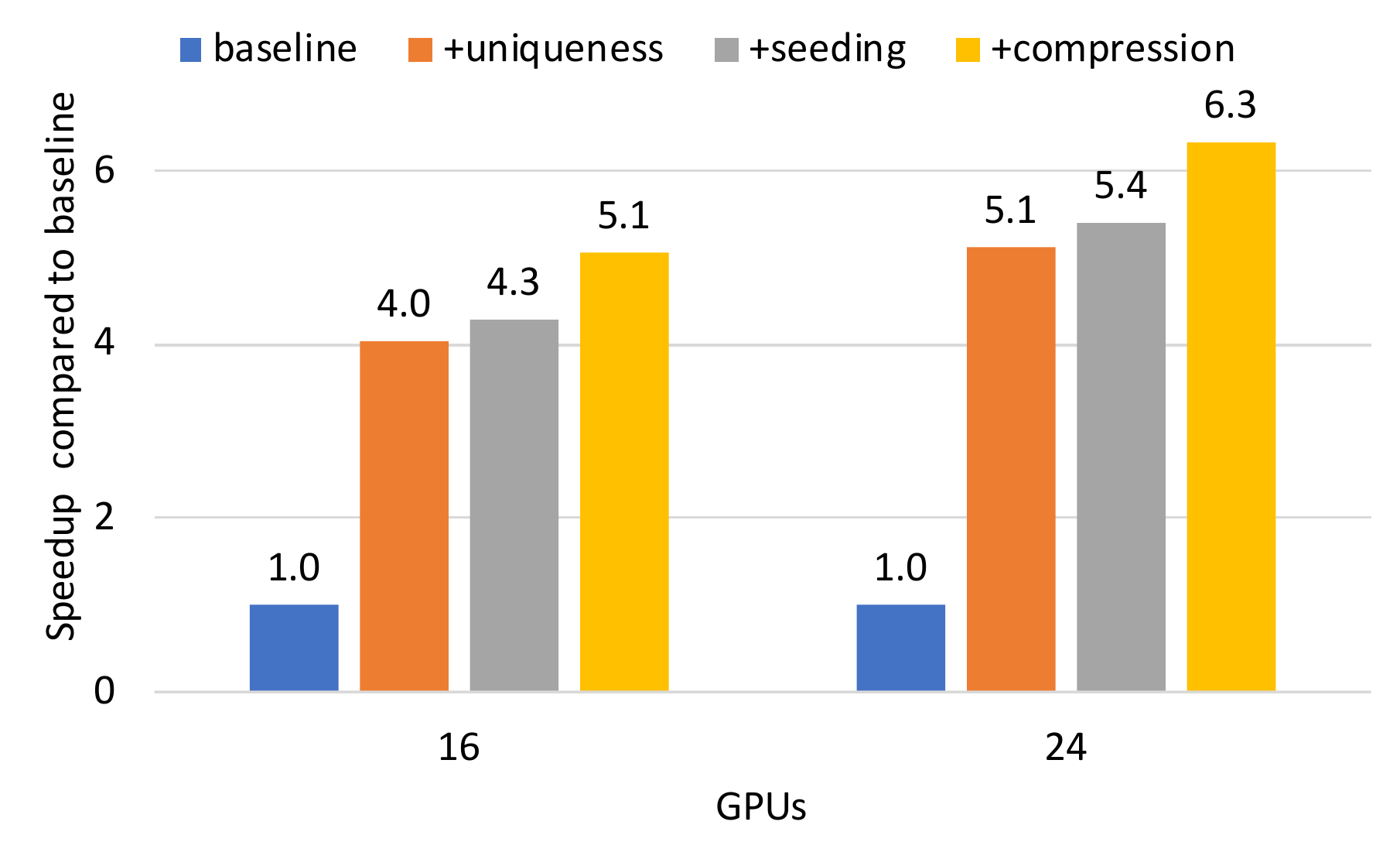}
  \redVspace
  \caption{Speedup achieved using our techniques, uniqueness, seeding, and compression (lower precision) compared to the baseline (without these techniques) word language model on 16 and 24 TitanXx8 GPUs.}
  \label{fig:speedup-baseline}
\end{center}  
\redVspace
\end{figure}

Figure \ref{fig:speedup-baseline} shows the performance improvement up by each of the three techniques---{\it uniqueness, seeding}, and {\it compression}.
To do this, we present the results obtained from using 16 and 24 GPUs on 1-Billion word dataset. We consider the baseline that does not use our techniques \cite{DBLP:journals/corr/abs-1712-00409}. 
{\it Uniqueness} delivers a 
4$\times$ performance improvement (speedup).
The speedup closely matches to the ratio of total and unique words  (Figure \ref{fig:statistics}), which is 3.4$\times$ at 16 GPUs. 
The {\it seeding} and {\it compression} techniques give additional 7\% and 18\% performance improvements, respectively, thus reaching a total of 5.1$\times$ speedup compared to the baseline. The speedup was found to be higher (e.g. 6.3$\times$ on 24 GPUs as shown in the Figure \ref{fig:speedup-baseline}) as the gap of unique words vs. total words increases with the number of GPUs. 
The peak GPU memory in use (not shown), without our techniques, grows linearly: 3.9 GB, 7.1 GB, and 10.3 GB per GPU at 8, 16, and 24 GPUs, respectively and goes out of memory after that. 
In contrast, the peak GPU memory in use, with our techniques, remains almost steady---1.19 GB at 8 GPUs, 1.20 GB
at 24 GPUs, and 1.21 GB at 64 GPUs.
Thus, we achieve 8.6$\times$ memory reduction when using 24 GPUs.

\begin{figure}[t!]
\begin{center}
  \includegraphics[width=0.5\textwidth]{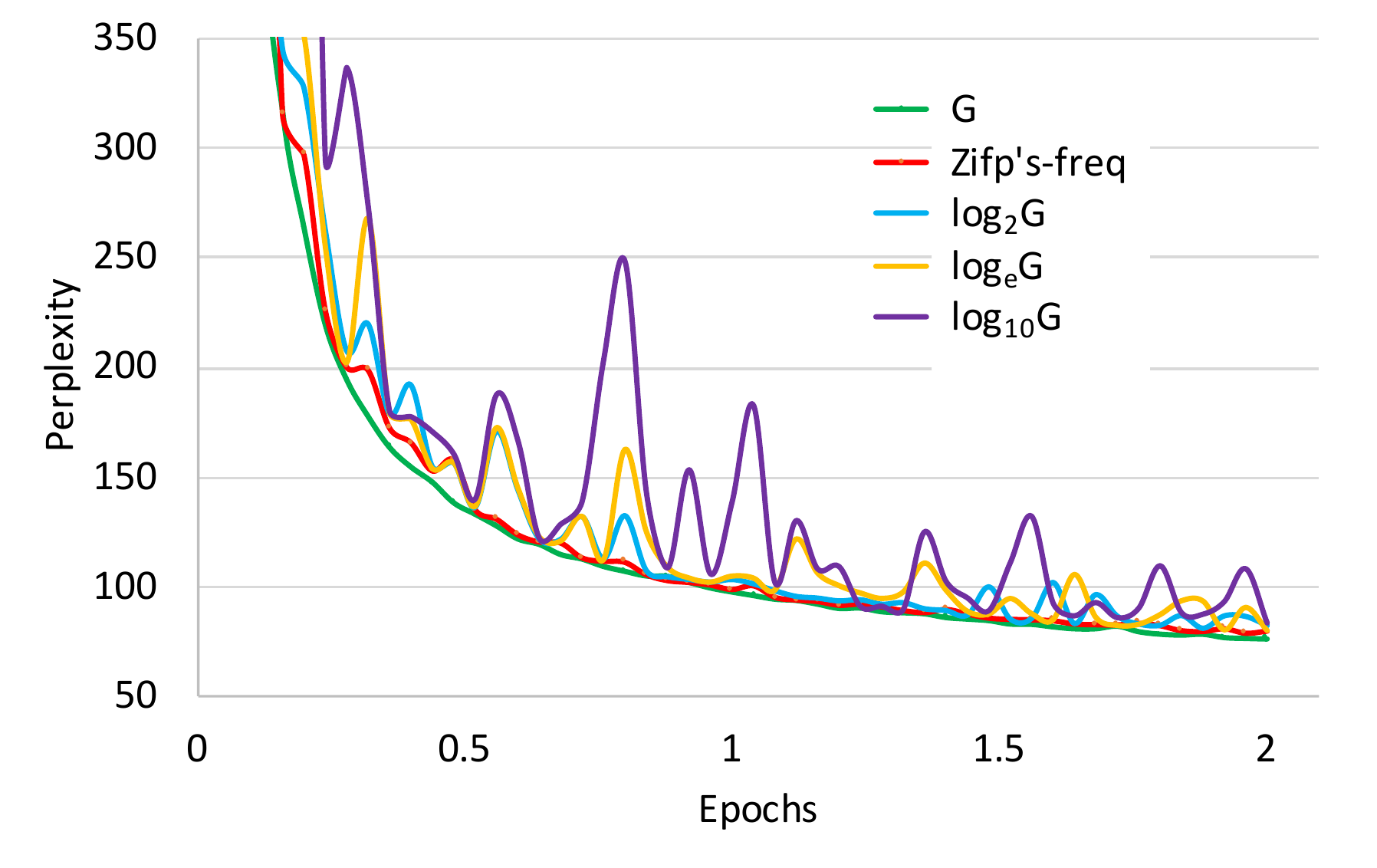}
  \redVspace
  \redVspace
  \caption{Different seeding techniques used in the sampled softmax layer for word language model using 64 GPUs.}
  \label{fig:seeding-techniques}
\end{center}  
\redVspace
\end{figure}

We now divert attention how our techniques may influence accuracy.
The {\it uniqueness} technique only changes the 
flow of computation as discussed in Section \ref{subsec:exploit-uniquness}, and hence produces the same accuracy
as the baseline for word language model. 

Figure \ref{fig:seeding-techniques} shows the impact of different seeding techniques on accuracy, which is used in the output
embedding layer to compute sampled softmax
for word language model. 
We used a different seed on each GPU (line with label \texttt{G}) and also the number of seeds equal to $\log_2$, $\log_e$, and $\log_{10}$ of the number of GPUs. We have 
also performed experiments where the number of seeds  
follows the word frequency distribution (line with label {\it Zipf's-freq}).
Decreasing the number of seeds makes the accuracy of the training curve less stable (e.g. $\log_2$ shows more close perplexity as \texttt{G} than $\log_{10}$). 
Seeding with {\it Zipf's-freq} produces similar perplexities as {\tt G} seeds and offers a pareto optimal setting. 

The {\it compression} technique loses 
lower precision bits, hence accuracy is expected to be lower. But {\it compression-scaling} (Section \ref{subsec:compression}) regains the same accuracy. 
For example, the perplexity of word language model 
after 1 epoch on 16 GPUs with and without compression are 84.12 and 84.68, respectively. 


\subsection{Character Language Model}
\label{subsec:char-LM-results}

\begin{figure}[t!]
\begin{center}
  \includegraphics[width=0.45\textwidth]{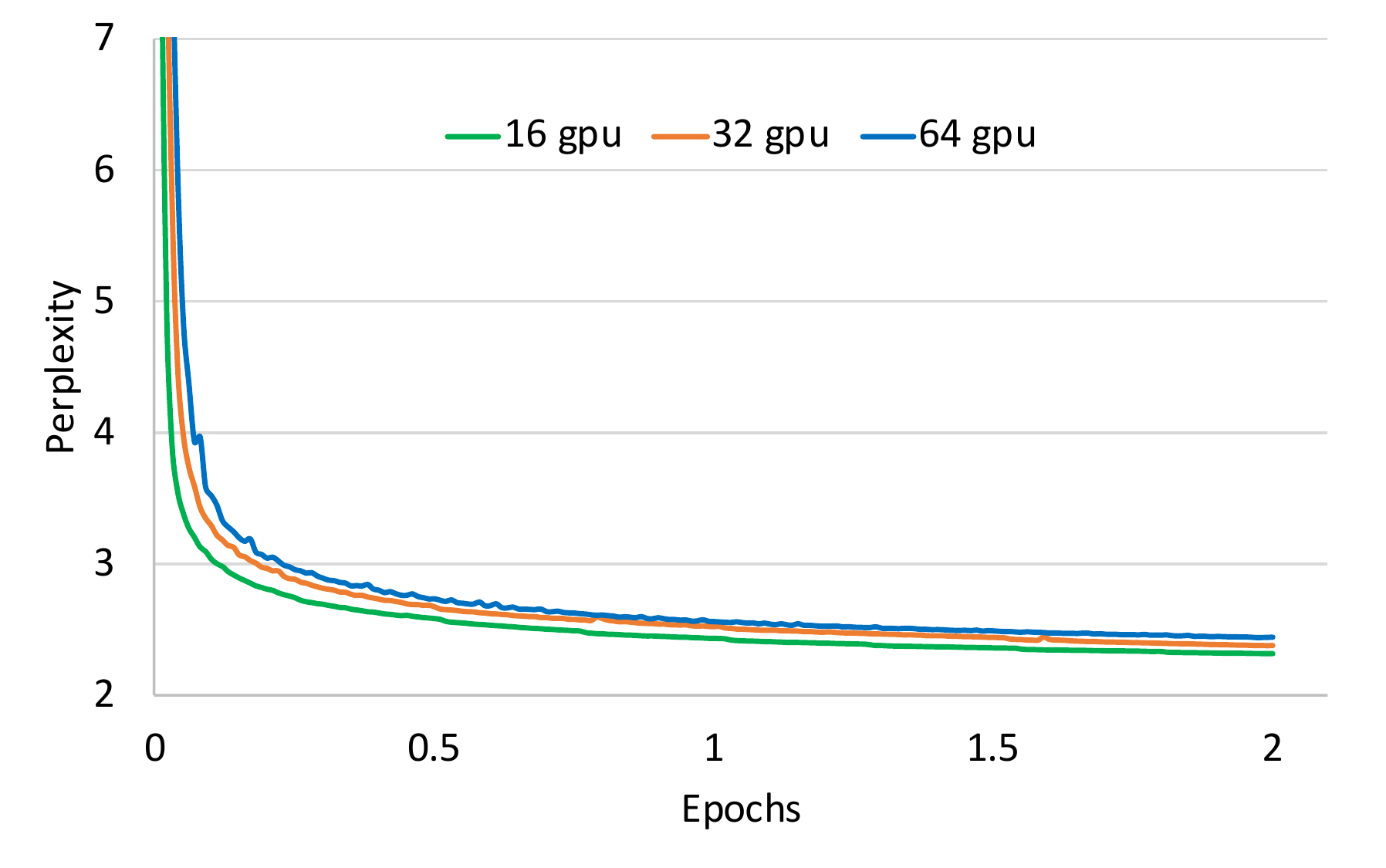}
  \redVspace
  \caption{Accuracy of character language model on the 1-Billion word dataset using 16, 32, and 64 GPUs.}
  \label{fig:charLM-accuracy-gpus}
\end{center}   
\redVspace
\end{figure}

Figure \ref{fig:charLM-accuracy-gpus} shows the accuracy (perplexity) up to 2 epochs for 
character language model with small vocabulary ($\sim100$) on the 1-Billion dataset. 
Similar to the word language model, we use
16, 32, and 64 GPUs, to perform the scaling experiments with a batch size of 2048, 4096, and 8192 (hence 0.3M, 
0.6M and 1.2M total characters), respectively. 
As the figure shows, our three sets of experiments produces
similar perplexities. We observe that gap of  perplexities reduces as we progress towards further epochs. For example, perplexity difference between 16 and 32 GPUs at epoch 1 is 4\%, whereas at epoch 2 and 4, the gap becomes 2\% and
0.01\%, respectively. We observe similar results when comparing 16 GPUs with 64 GPUs (the gap is 5\% at epoch 1 and 1\% at epoch 5). Although the perplexity with higher GPUs has higher perplexity at any point in the figure, running a few additional iterations produces the same accuracy as the lower number of GPUs (e.g. perplexity of 2.27 using 16 and 32 GPUs at epoch 3 and 3.4, respectively). 

We observe similar results on the Gutenberg dataset. At epoch 1, the perplexity of 16 and 32 GPUs are 2.78 and 2.85, respectively. However, at epoch 3, the corresponding values become 2.53 and 2.54. 
Similar results have been observed when comparing the accuracy of 16 vs. 64 GPUs. Note that 
we increased the base learning learning rate ($10^{-3}$ 
for 8 GPUs) by a multiplying factor of $\log_e |nodes|$ 
(e.g. $2.07\times 10^{-3}$ for 64 GPUs), similar to the word LM.

We now discuss how the training time per epoch for the 1-Billion word dataset reduces
as we increase the number of GPUs, while keeping the local batch 
size fixed. Table \ref{tab:epoch_time_char_gpus} shows the taken 
time and parallel efficiency with and without our techniques. 
We use the runtime using 8 GPUs as the baseline for comparison among the experiments. 
Our techniques take 23.2 hours per epoch using 8 GPUs and increasing the GPUs to 64, the time reduces to 3.5 hours. We achieve 6.6$\times$ speedup (with 82\% parallel efficiency) using 8$\times$ more GPUs.
At 24 GPUs, while our technique delivers 94\% parallel efficiency, without our techniques, the baseline delivers 81\% parallel efficiency.
Beyond 24 GPUs, the baseline goes out of memory, whereas our implementation continues to scale---a demonstration of the usefulness of our \emph{uniqueness} and \emph{compression} techniques detailed in Section \ref{sec:chall-techn}. 
Note that \emph{seeding} technique was not used for character 
LM as the vocabulary size is small, hence
full softmax was used instead of sampled softmax layer.
We achieved similar speedup (6.7$\times$ using 8$\times$ GPUs compared to 8 GPUs baseline) when we experiment on the Gutenberg dataset. We obtained 3.95 TFLOP/sec (64\% of peak FLOPS) in the character LM experiments.
We mention in passing that the number of unique \emph{characters} becomes constant (reaching the size of the small vocabulary) as we keep increasing the batch size (thus GPUs) in character language model.

\begin{table}[t!]
\scriptsize
\begin{center}
\caption{Per epoch time  (hours) on Titan X GPUs for character LM using 1-Billion word dataset. 8-GPU is the  baseline to compute parallel efficiency. $*=>$ out of GPU memory.}
\label{tab:epoch_time_char_gpus}
\begin{tabular}{c||c|c||c|c} \hline
\multicolumn{1}{c||}{} & \multicolumn{2}{c||}{Without Our Technique} & \multicolumn{2}{c}{With Our Technique} \\  \hline
GPUs & Time & Parallel Efficiency & Time & Parallel Efficiency \\ \hline 
8 &	25.7	& 100\% &	23.2 & 100\%	\\ 
16 &	14.5	& 89\% &	12.9 & 96\%	\\
24 &	10.6	& 81\% &	8.2 & 94\%	\\ 
32 	&	*	& - &	6.8 & 86\%	\\
64 	&	*	& - &	3.5 & 82\%	\\ \hline
\end{tabular}
\end{center}
\redVspace
\end{table}

 Table \ref{tab:epoch_time_char_gpus} shows an improvement in performance 
when compared to the same number of GPUs without our techniques.
For example, on 16 GPUs, we found {\it uniqueness} contributes to 23\% runtime reduction. 
We observe limited gain (e.g. 2\% on 16 GPUs) using the {\it compression} technique for character LM.
This is mainly due to the fact that the character language model has higher number of tensors ($> 20$), each 
needs to down-cast (FP32 $\rightarrow$ FP16) and up-cast (FP16 $\rightarrow$ FP32), thus 
adds an overhead to get benefit of the compression technique. When we compared the accuracy, 
we found our {\it compression-scaling} (Section \ref{subsec:compression}) technique 
regains the same accuracy as without using compression.
For example, the perplexity of character language model 
after 1 epoch on 64 GPUs with and without compression are 2.58 and 2.59, respectively.


\subsection{Hero Scale Run (Tieba dataset, 192 GPUs)}
\label{subsec:hero-scale-runs}

In this section, we apply our techniques to train massive data that was impractical previously. 
We improve the accuracy of language modeling on the Tieba \cite{baidutieba2018} dataset,  keeping the training time in a reasonable range while scaling to more GPUs and hence training on more data.

We take two subsets, 1 and 4 Billion Chinese characters  from the Tieba dataset \cite{baidutieba2018} (32 Billion). 
We use the same validation set to test accuracy of all three datasets. 
The vocabulary we used consists of 15,437 characters ($\sim\!\!\!150\times$ larger than English, thus
a demonstration of scaling character language model with large vocabulary).
We perform weak scaling using 6, 24, and 192 GPUs for the 1B, 4B, and 32B datasets respectively.
The corresponding learning rate is $2 \times 10^{-4}$, $4 \times 10^{-4}$, and $5 \times 10^{-4}$. 
Table \ref{tab:tieba-results} shows that increasing the data size from 1B by 4$\times$ and 32$\times$,
the training taken time per epoch increases by only 1.04$\times$ and 1.25$\times$, respectively. We achieve a total of 0.76 PFLOP/s using 192 GPUs.
Compared to 6 GPUs with 3GB corpus, a 12 GB corpus on 24 GPUs delivers a 20\% accuracy improvement and a 93 GB corpus on 192 GPUs delivers 35\% accuracy improvement.

Since the internal Tieba dataset does not have public baselines on accuracy, we compute
the compression ratio as a metric to demonstrate the competitiveness of our results on this corpus.
We chose this metric as perplexity is an indication of performance in text compression.
We compute the compression ratio by dividing the corpus size by 
the product of bits per character and total number of characters in the corpus.
\cite{2018arXiv180801371P} showed a
bit per character (i.e. $\log_2(\text {perplexity})$) of 1.11 for the 
Amazon review dataset with comparable batch size, 
which equates to a compression ratio of
6.8. 
For the Tieba dataset (93GB, 34 Billion Characters), 
we achieve a comparable compression ratio (e.g. the perplexity of 11.1 equates 
to compression ratio of 6.3).


\subsection{Comparison with the Existing Results}
\label{subsec:compare-sota}

We compare our results with a recent work on scaling language  modeling  \cite{2018arXiv180801371P},
despite the fact that our implementation is capable of scaling on more GPUs and 
larger vocabularies (i.e. 192 GPUs, 15K and 100K vocabulary for character and word LM, respectively) 
than \cite{2018arXiv180801371P} (128 GPUs and small
vocabulary of 100). 
Although the dataset they used in the experiments is publicly available (e.g. Amazon review 
\cite{McAuley:2015:IRS:2766462.2767755}), the infrastructure is the most recent one (October, 2018) and 
specialized. For example, the 128 GPUs used were V100 (peak 125 TFLOP/s, 16GB of HBM2 memory,
and NVLink to communicate among GPUs). Since we do not have access to such infrastructure,
we perform experiments using 64 Titan X GPUs (peak 6.1 TFLOP/s, 12GB of HBM2 memory, and PCIe 
for communication). Using the above discussed RHN based character LM, 
we achieve an accuracy of 1.208 BPC (bit per character) compared
to 1.218 reported in \cite{2018arXiv180801371P} after 1 epoch. 
When compared the training time, we take 17.6 hours, 14$\times$ longer than 
\cite{2018arXiv180801371P}, but using 41X less powerful infrastructure 
(16 PFLOP/s vs. 0.39 PFLOP/s), leading to a rough gain of 2.9$\times$. The gain increases to 3.3$\times$
as we train to 3 epochs with an accuracy of 1.11 BPC.




\begin{table}[t!]
\scriptsize
\begin{center}
\caption{Tieba results.}
\label{tab:tieba-results}
\begin{tabular}{c|c|c|c|c|c} \hline
Characters & Corpus & GPUs & Batch & Time & Perplexity\\
(Billion) & (GB) &  & Size & (hours) & (1 epoch) \\  \hline 
1.07  &  3 & 6 			& 768 &	27 & 17.06 \\  
4.29 	&	12 & 24	& 3,072 &	28 & 13.6 \\  
34.36 &	93 &	192	& 12,288 &	34 & 11.1 \\ \hline 
\end{tabular}
\end{center}
\redVspace
\end{table}






\section{Related Work} 
\label{sec:related_works}

Compute required to train deep neural networks jumped 15$\times$ and compute delivered by GPUs increased by 10$\times$, just in 2 years, 2015-2017 \cite{compute-need-delivered-2017}. Large-scale training has been of significant interest to reduce the training time. Most of the recent scaling efforts are centered around vision applications, such as image recognition and segmentation. For example, \cite{DBLP:journals/corr/GoyalDGNWKTJH17} trains ResNet-50 model using ImageNet dataset (1.2 million images) 
\cite{imagenet_cvpr09} in an hour using 256 Tesla P100 GPUs.
\cite{DBLP:journals/corr/GoyalDGNWKTJH17} reduces the training time to 
20 minutes using 2048 Intel Xeon Phi coprocessors.
\cite{DBLP:journals/corr/abs-1711-04325} goes further reducing
the training time to 15 minutes using 1024 Tesla P100 GPUs.

The importance of scaling has also been realized in the neural language
processing (NLP) domain, specially in language modeling, which plays a 
key role in traditional NLP tasks \cite{DBLP:journals/corr/JozefowiczVSSW16}.
For example, \cite{DBLP:journals/corr/JozefowiczVSSW16} performs experiments on a wide range of RNN based models and proposed a CNN based softmax loss computation, which improves accuracy on 1-Billion word dataset. The paper uses 32 Tesla K40 GPUs with asynchronous gradient updates. 
However, it has been shown that synchronous SGD can often converge to a better final accuracy than asynchronous SGD \cite{chen2016revisiting}. 
Moreover, asynchrony could effectively increase the momentum which is part of why it tends to diverge so easily \cite{mitliagkas2016asynchrony,jin2016scale}.
\cite{DBLP:journals/corr/abs-1804-03235} explores an online distillation-based large-scale distributed training method. The paper showes that codistillation works well on a wide range of applications including language modeling using 128 GPUs. But in the distillation approach, multiple models are trained in parallel, which significantly increases computation. \cite{DBLP:journals/corr/abs-1803-08240}
 scales both on word and character language model using
eight NVIDIA Volta GPUs. The dataset for character LM were $\sim\!90M$ and for word LM, it was $\sim\!100M$.  \cite{2018arXiv180801371P} scales character LM (small vocabulary of 100) using up to 128 NVIDIA Volta GPU using mixed precision training on 40 GB of Amazon review dataset. 



\section{Conclusions} 
\label{sec:conclusions}

Language modeling is a central problem in
natural language processing, which is used in many applications such as speech recognition and machine translation.
Prior work on language modeling has achieved limited scalability.
The \ag{} operations performed in the input and output \emph{embedding} layers of language models require large memory footprint which quickly grows out of GPU memory limits and demand large volume data exchange among GPUs.
In this paper, we showed how Zipf's law can be used to 
reduce the asymptotic complexity of both memory (within a GPU) and communication (across GPUs) and hence scale up language modeling to take advantage of more training data and more GPUs.
Using several datasets, we demonstrate 6.7$\times$ (character LM) and 6.3$\times$ (word LM) speedup by scaling to 8$\times$ more GPUs with negligible
loss of accuracy. 
Finally, we weak scale LM from six to 192 GPUs, which allows us to scale training from 3GB to 93GB of the Chinese Tieba dataset while taking only 1.25$\times$ more training time.
This weak scaling delivers 35\% more accuracy in predictions.

\end{document}